\documentclass[sigconf]{acmart}

\usepackage[table]{xcolor}
\usepackage{multirow}
\usepackage{amsmath}
\usepackage{wrapfig}
\usepackage{booktabs}  
\usepackage{caption}   
\usepackage{subcaption}
\usepackage{arydshln}
\usepackage{bbding}
\usepackage{hyperref}
\usepackage{float} 
\usepackage{makecell}
\hypersetup{
    colorlinks=true,    
    linkcolor=black,     
    urlcolor=blue,      
    citecolor=red      
}

\definecolor{lightpurple}{RGB}{240, 230, 250}
\definecolor{lightblue}{RGB}{224,242,247}
\definecolor{customgreen}{RGB}{0, 125, 3}
\definecolor{rowgreen}{RGB}{213,232,212}
\definecolor{darkgraycolor}{RGB}{120, 133, 137}

\AtBeginDocument{%
  }

\setcopyright{acmlicensed}
\copyrightyear{2026}
\acmYear{2026}
\setcopyright{cc}
\setcctype{by}
\acmConference[DAC '26]{63rd ACM/IEEE Design Automation Conference}{July 26--29, 2026}{Long Beach, CA, USA}
\acmBooktitle{63rd ACM/IEEE Design Automation Conference (DAC '26), July 26--29, 2026, Long Beach, CA, USA}
\acmDOI{10.1145/3770743.3804012}
\acmISBN{979-8-4007-2254-7/2026/07}

\begin{document}

\title{LiveVLM: Efficient Online Video Understanding via Streaming-Oriented KV Cache and Retrieval}


\author{
  Zhenyu Ning$^{1}$, 
  Guangda Liu$^{1}$, 
  Qihao Jin$^{2}$, 
  Chengwei Li$^{1}$, 
  Wenchao Ding$^{2}$, 
  Minyi Guo$^{3,1}$, 
  Jieru Zhao$^{1\dag}$ \\
}
\affiliation{%
  \country{$^{1}$School of Computer Science, Shanghai Jiao Tong University}
}
\affiliation{%
  \country{$^{2}$College of Intelligent Robotics and Advanced Manufacturing, Fudan University}
}
\affiliation{%
  \country{$^{3}$Guizhou Provincial Laboratory of Big Data, College of Computer Science and Technology, Guizhou University}
}
\affiliation{%
  \country{$^\dag$Corresponding author: zhao-jieru@sjtu.edu.cn}
}

\renewcommand{\shortauthors}{Ning et al.}

\begin{abstract}
Recent developments in Video Large Language Models (Video LLMs) have enabled models to process hour-long videos and exhibit exceptional performance.
Nonetheless, the Key-Value (KV) cache expands linearly over time, leading to substantial memory overhead and response delay--critical challenges in various real-world online applications, such as Deepseek services, autonomous driving and robotics. 
To mitigate these issues, we propose \textbf{LiveVLM}, a training-free and query-agnostic framework specifically designed for online video understanding and real-time interaction. 
LiveVLM employs a Vision Sink Bucketing (VSB) mechanism to process video streams in real time, retain long-term video details and eliminate redundant KVs. 
This mechanism utilizes vision-to-vision attention scores as the metric and seeks to maximize the coverage of contextual information during compression.
Noting that KV cache compressed in a query-agnostic manner inevitably retains irrelevant information for specific queries, LiveVLM incorporates a Position-agnostic KV Retrieval (PaR) mechanism to reduce interference from redundant context.
The keypoint of PaR lies in decoupling positional embeddings to enhance the similarity between key tensors, thereby supporting efficient retrieval at the granularity of pages. 
Extensive experiments demonstrate that LiveVLM enables the foundation LLaVA-OneVision model to achieve state-of-the-art accuracy among both training-free query-agnostic methods and training-based online models. Our code is available at \url{https://github.com/sjtu-zhao-lab/LiveVLM}.

\end{abstract}

\maketitle
\section{Introduction}

\begin{figure}[htbp] 
    \centering
    \begin{subfigure}[b]{\linewidth} 
        \centering
        \includegraphics[width=0.95\linewidth]{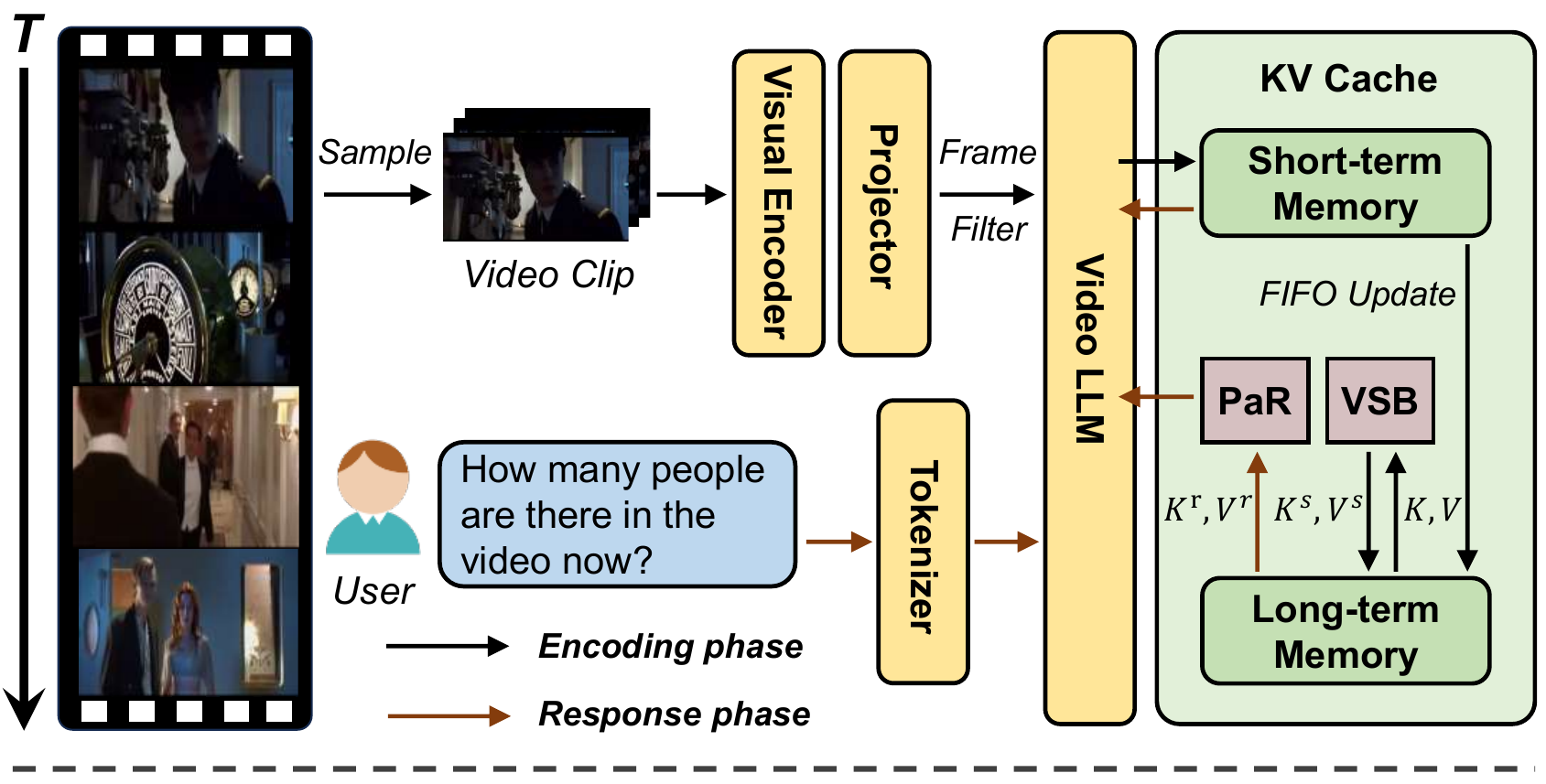} 
    \end{subfigure}
    \begin{subfigure}[b]{0.49\linewidth} 
        \centering
        \includegraphics[width=\linewidth]{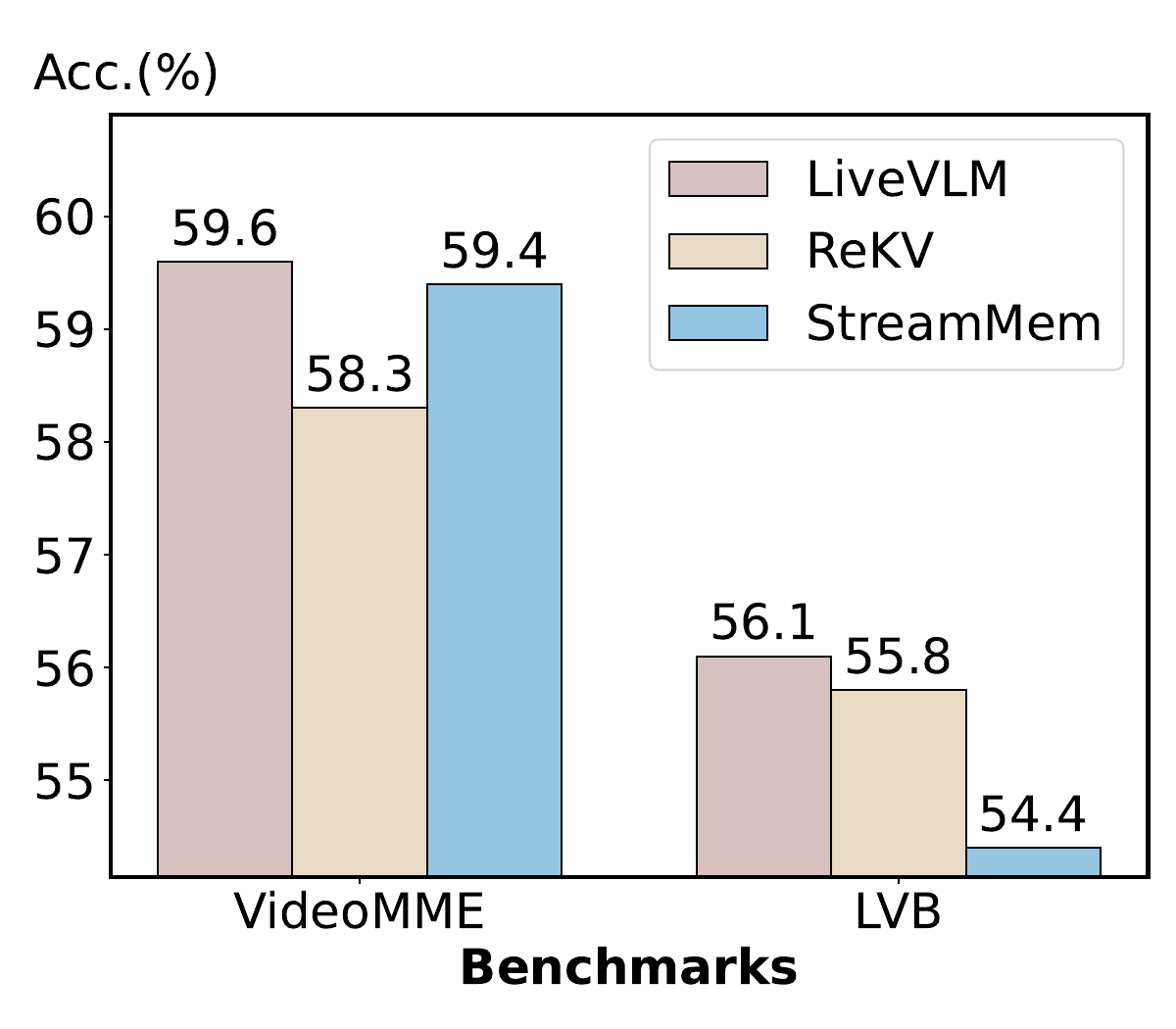} 
        \caption{Performance}
    \end{subfigure}
    \begin{subfigure}[b]{0.49\linewidth} 
        \centering
        \includegraphics[width=\linewidth]{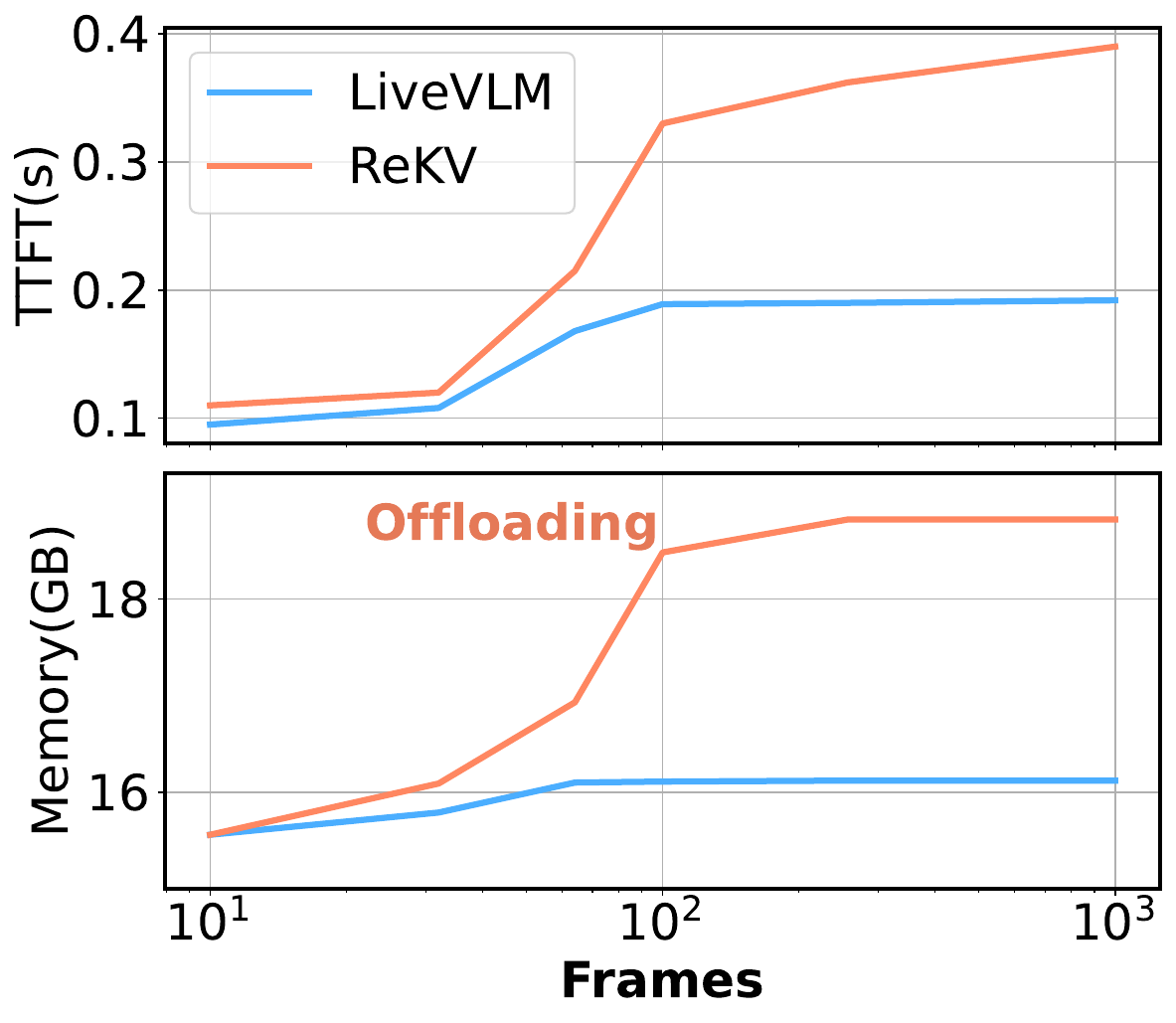} 
        \caption{Speed \& Memory}
    \end{subfigure}
    \vspace{-0.3cm}
    \caption{The overview of LiveVLM.} 
    \label{fig:overview} 
    \vspace{-0.5cm}
\end{figure}

Building upon the advancements in LLMs \cite{gemini,gpt-4o,deepseek}, Video Large Language Models (Video LLMs) \cite{gpt4v,qwen2-vl,qwen2.5-vl} have been introduced to facilitate applications such as autonomous driving, robotics \cite{embodiedgpt, wang2024large} and augmented reality (AR) devices. 
To reduce training overhead and enable multimodal rationalization, Video LLMs encode videos into visual features using a pre-trained visual encoder \cite{ve1,ve3} and leverage a projection module \cite{blip2} to align these features with text tokens from the tokenizer \cite{llava}. 
These visual features, also called vision tokens, are then input to the LLM along with text tokens.
Current Video LLMs have achieved exceptional performance across various tasks, such as video captioning and action recognition. 
Most Video LLMs are designed in an offline setting: the entire video is input to the model once a question is posed. However, the processing becomes much more challenging in 
online scenarios, where models are required to perform continuous processing of video streams before a query arrives (\textit{encoding phase}) and generate responses 
upon receiving new queries in real time (\textit{response phase}). The challenges persist in three aspects: \textbf{accurate video understanding (\textit{quality})}, \textbf{large memory usage (\textit{overhead})} and \textbf{real-time response (\textit{speed})}.


Prior studies attempt to reduce memory consumption while preserving accuracy and processing speed. 
Query-dependent compression methods \cite{moviechat+,llamavid} compress video contents based on the current query by retaining only query-relevant tokens, thereby achieving high compression ratios. 
However, these approaches often suffer from accuracy degradation because vision tokens that are relevant for future queries can be prematurely discarded.
Moreover, their processing pipeline is triggered on after a new query arrives, leaving the system idle during the video encoding phase and ultimately leading to high response latency.
To resolve this issue, some methods prune redundant vision tokens 
query-agnostically, using strategies such as sampling \cite{tang2025adaptive}, pooling \cite{llavaonevision, videollm-online} and fusion \cite{ma_lmm, vstream}. However, for each new query, these methods still incur $O(n^2)$ prefill computation complexity over all vision tokens during response phase. This forces them to adopt extremely aggressive compression to meet memory and latency constraints, degrading the response accuracy. 

Recent works have proposed query-agnostic KV cache compression approaches. By computing KVs as video frames arrive, this paradigm reduces the computational complexity during response phase, thereby achieving superior speed compared with vision token pruning methods.
To mitigate memory overhead, these methods either offload KVs to the CPU \cite{rekv} or permanently discard KVs \cite{infinipotv, streammem}. 
Although retaining full Key-Value (KV) cache in the CPU and retrieving them based on queries facilitates the preservation and retrieval of fine-grained information, it leads to severe inefficiency. In contrast, while permanent discarding achieves low memory overhead and real-time performance, this approach exhibits inferior performance in tasks that require long-term temporal details. 
In conclusion, existing methods fail to simultaneously address the three aforementioned challenges, namely quality, memory overhead, and response speed.

\begin{figure}[t]
  \centering
  \includegraphics[width=0.79\linewidth]{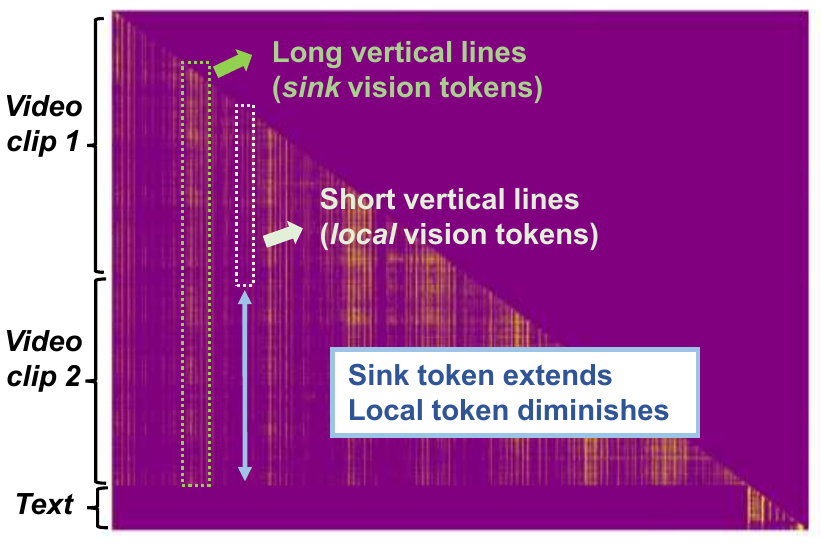}
  \vspace{-0.26cm}
  \caption{Visualization of sparse attention patterns, encompassing both vision-to-vision and text-to-vision patterns.}
  \label{fig:attn_pattern}
  \vspace{-0.4cm}
\end{figure}

To this end, we propose LiveVLM, a training-free and query-agnostic framework designed for online video understanding and real-time interaction (Fig. \ref{fig:overview}).
During encoding phase, the framework efficiently encodes and stores online video streams via KV generation and compression.
LiveVLM employs a novel \textit{\textbf{Vision Sink Bucketing (VSB)}} compression mechanism.
VSB leverages our key observation (Fig. \ref{fig:attn_pattern}) of \textit{sink} vision tokens--tokens that consistently attract high attention from others, as retaining them maintains the attention outputs of newly arriving tokens close to pre-compression levels. 
Notably, VSB identifies sink vision tokens by vision-to-vision attention scores, which differs from mainstream KV cache compression approaches that rely on text-to-vision attention scores.
We further observe that selecting tokens solely by attention scores retains redundant \textit{local} tokens (locally focused and neglected by subsequent tokens). 
Building on these insights, VSB retains tokens by considering both attention scores and maximizing context coverage for long-term information. 

When a new question arrives, LiveVLM retrieves relevant KVs from the cache and feeds them with text tokens into the LLM to generate an answer. By leveraging pre-computed KVs, LiveVLM reduces the attention computation to interactions between query tensors of text tokens and retrieved KVs, lowering computational complexity from $O(n^2)$ to approximately $O(n)$ and significantly accelerating response speed. To ensure accurate and timely responses, we introduce an \textit{\textbf{Position-agnostic KV Retrieval (PaR)}} mechanism that efficiently fetches long-term information (relevant KVs) from the cache based on the current query, while maintaining a sliding window of KVs from the most recent video clips to capture detailed short-term context. 
Both long-term and short-term KVs are then provided to the LLM alongside the input query for final computation. 
The novelty of PaR resides in decoupling positional embeddings to augment the similarity between key tensors, thereby facilitating efficient retrieval at the page-level granularity. 

\begin{figure}[t]
  \centering
  \includegraphics[width=0.87\linewidth]{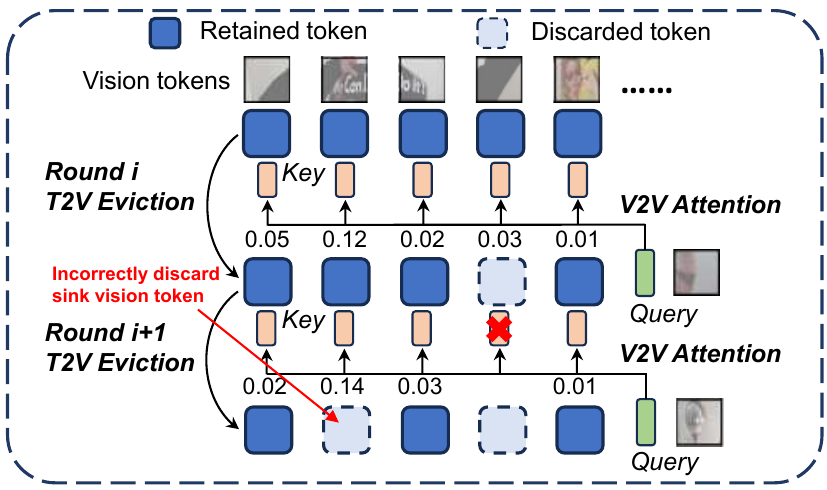}
  \vspace{-0.2cm}
  \caption{Distinctions between text-to-vision (T2V) and vision-to-vision (V2V) attention distribution. 
  }
  \label{fig:round_attn}
  \vspace{-0.3cm}
\end{figure}

\section{Related Work}
\subsection{Video Large Language Models}
The notable advancement in LLMs has driven the development of Video Large Language Models (Video LLMs) \cite{llava, video_chatgpt, video-llava}. 
These models employ a visual encoder for video feature extraction, aligning such features with text tokens through Linear Projection, MLP, or Q-Former.
The aligned visual features are then combined with text tokens and input to backbone LLMs.
Though achieving strong performance on short video understanding benchmarks, they encounter challenges in long video understanding or streaming video understanding due to the memory bottleneck and the information complexity of long videos.
To address these challenges, many offline strategies have been proposed, including sparse sampling \cite{pllava}, instruction-guided token compression, and token merging \cite{chatunivi, slowfast}.
However, these strategies often incur substantial information loss or inefficiencies.

\subsection{KV Cache Compression for Video LLMs}
KV cache compression techniques are widely implemented in both LLMs and Video LLMs as an effective means to enhance inference speed and reduce memory consumption \cite{fastv, dycoke, mminference}. 
These methods leverage the inherent sparsity of the attention computation and discard tokens with low attention scores.
FastV \cite{fastv} reveals that vision tokens generally exhibit significantly lower attention scores than text tokens, reducing vision tokens by over 50\% in the decode phase while maintaining accuracy.
Recent MMInference \cite{mminference} further explores unimodal and cross-modal sparse patterns to accelerate prefill computation.
However, most existing methods depend on query-derived text tokens to compute attention scores and initiate compression, inevitably causing response delays in online scenarios.
To address this challenge, LiveVLM explores the sparse attention patterns of vision tokens and designs a streaming-oriented cache compression strategy.

\begin{figure}[t]
  \centering
  \includegraphics[width=0.97\linewidth]{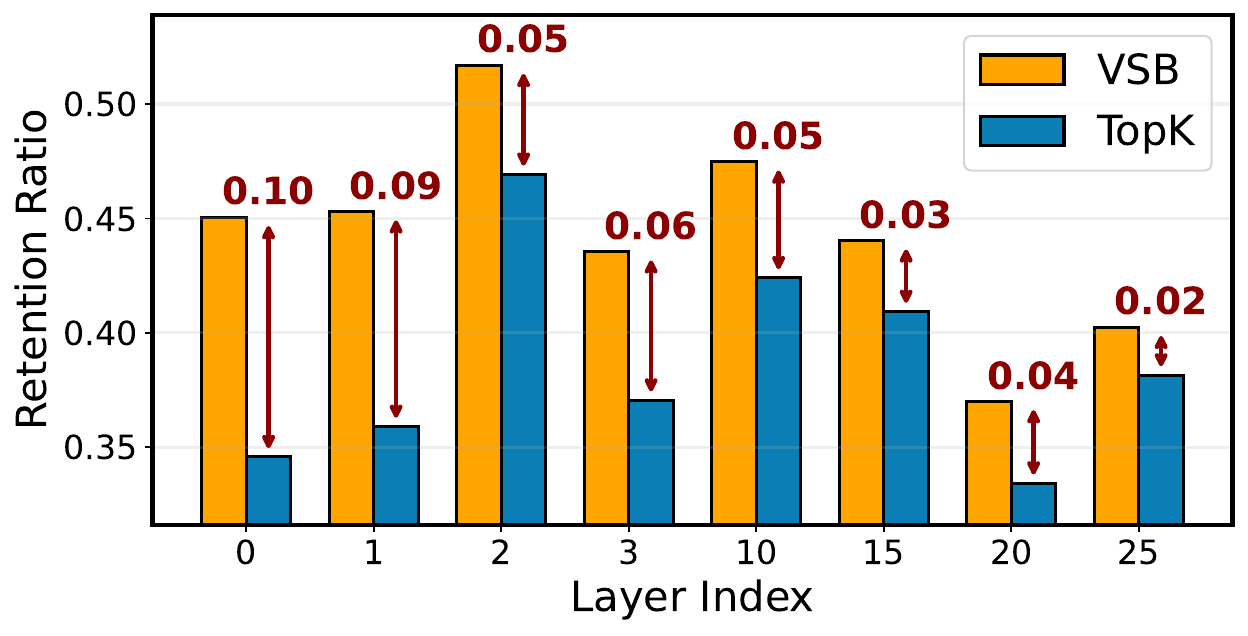}
  \vspace{-0.4cm}
  \caption{Comparison of retention ratio between TopK and VSB. 
  }
  \label{fig:overlap_bar}
  \vspace{-0.4cm}
\end{figure}

\subsection{Streaming Video Understanding}
Streaming video understanding requires Video LLMs to process continuously updating video frames and respond to user queries based on content from the start to a specified timestamp.
VideoLLM-online \cite{videollm-online} is a leading work enabling efficient processing and proactive response, but its pooling strategy significantly degrades spatial details.
Subsequent studies exemplified by Flash-VStream \cite{vstream} and Dispider \cite{dispider} organize visual features via dedicated memory-augmented architectures, which are concatenated with text tokens and input to models in an offline manner.
Recent works have proposed a query-agnostic compression paradigm. 
ReKV \cite{rekv} offloads the full cache to the CPU for storage and conducts in-context retrieval of relevant KVs during query answering.
Nevertheless, this method entails considerable memory consumption and response latency when processing long-duration videos. 
In contrast, InfiniPot-V \cite{infinipotv} and StreamMem \cite{streammem} adhere to a constrained memory budget through diverse compression techniques, including temporal-axis redundancy reduction, value norm-based selection, and template-based selection.
Although these approaches perform effectively in general tasks, they suffer from long-term information loss and fail to handle specific tasks.
LiveVLM resolves this issue via its VSB strategy, which preserves global contextual information under a moderate cache budget, and employs PaR to capture visual details and enhance response efficiency.

\section{Vision Sink Bucketing}

\subsection{Sparse Attention Patterns of Vision Tokens}

We visualize attention patterns in Fig. \ref{fig:attn_pattern}. As illustrated, text-to-vision and vision-to-vision attention patterns exhibit significant discrepancies.
Text tokens pay considerable attention to other text tokens while allocating negligible attention to vision tokens. 
In contrast, vision tokens demonstrate complex inter-token attention patterns, with \textit{sink} vision tokens manifested as long vertical lines in the figure being a prominent feature. 
These sink vision tokens sustain high attention scores from other vision tokens, thus exerting a significant impact on the results of attention computation. Consequently, retaining such tokens helps alleviate compression-induced deviations in streaming inference.
In contrast, text tokens lack the capability to identify sink vision tokens and tend to discard them in the compression process (as depicted in Fig. \ref{fig:round_attn}), thereby disrupting the attention computation outcomes of subsequent vision tokens. 
As the video stream progresses, such deviations accumulate incrementally, ultimately resulting in the loss of video information and degraded model performance.
Therefore, we use vision-to-vision attention scores 
as the importance metric to retain critical vision tokens in online scenarios.

\begin{figure}[t]
  \centering
  \includegraphics[width=0.98\linewidth]{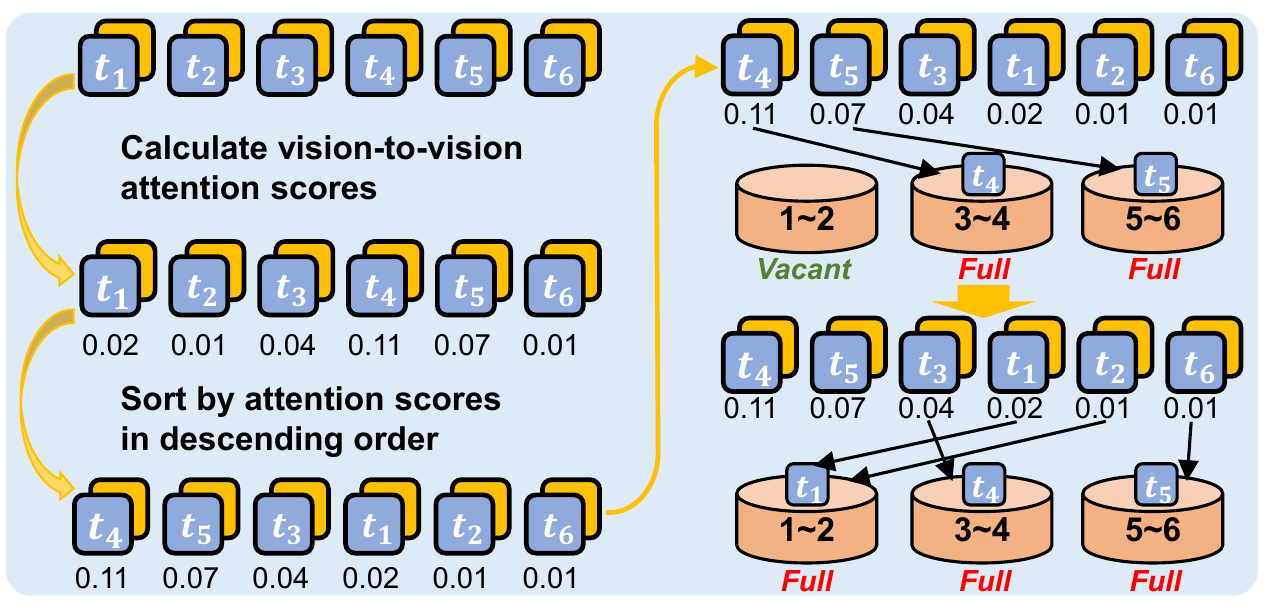}
  \vspace{-0.2cm}
  \caption{Illustration of the Vision Sink Bucketing (VSB) strategy. 
  }
  \label{fig:vsb}
  \vspace{-1em}
\end{figure}

Existing compression methods based on attention scores typically select and retain KVs corresponding to the top-$M$ tokens with highest scores from the current KV cache as the compressed cache.
Here, $M$ denotes the predefined upper limit of the cache size. 
Following this paradigm, we replace the text-to-vision attention scores with vision-to-vision attention scores.
However, based on our observations, such compression approaches would retain \textit{local} vision tokens that receive high attention from adjacent vision tokens yet fail to sustain attention from subsequently arriving tokens.
Retaining these redundant tokens impairs compression performance and hinder the cache from accommodating new critical vision tokens. 
Therefore, we propose our VSB strategy to balance the sum of attention scores and the covered context range of selected vision tokens.

We conduct an experiment to directly retrieve answer tokens from the complete context using the specific question tokens, and compare the results with two alternative approaches: the streaming compression method that exclusively relies on attention scores (termed ``TopK'') and our proposed VSB strategy.  
The experimental results are presented in Fig.\ref{fig:overlap_bar}.
The retention ratio refers to the ratio of answer tokens retained by the compression method to the total number of answer tokens during streaming inference. 
The results demonstrate that the VSB strategy can retain more potential answer tokens across all model layers (especially shallow layers), thereby better adapting to query-agnostic online scenarios. 

\begin{figure}[t]
  \centering
  \includegraphics[width=0.91\linewidth]{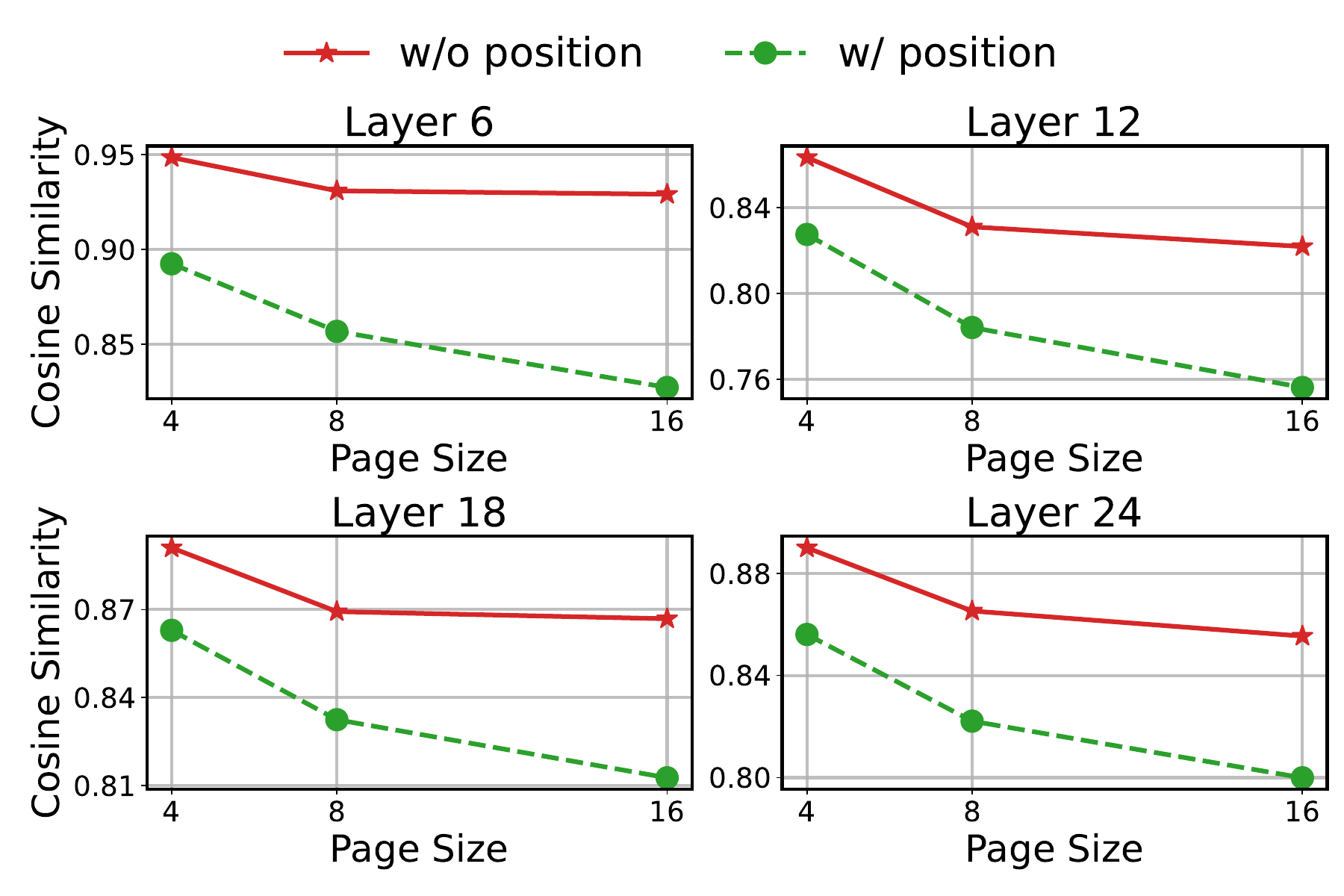}
  \vspace{-0.4cm}
  \caption{Comparison of cosine similarity between retrieval with and without positional embeddings.}
  \label{fig:layer_cosine_similarity}
  \vspace{-0.2cm}
\end{figure}

\subsection{Algorithm Design} 
\label{sec:svb}

In this section, we introduce the algorithm design of Vision Sink Bucketing (VSB) strategy (as illustrated in Fig.\ref{fig:vsb}). Firstly, it is necessary to obtain the vision-to-vision attention scores.
Since latest Video LLMs are integrated with FlashAttention \cite{dao2022flashattention}, the attention scores are not explicitly calculated. 
To address this, VSB computes partial attention scores by adopting the last $r$ vision tokens ($r\ll L$, where $L$ denotes the context length) as an observation window \cite{snapkv}, which is enough to capture sink vision tokens essential to our strategy.
Further, it derives the mean importance scores $\mathbf{S}$, formulated as follows:
\begin{align*}
    \mathbf{W} &= \text{softmax}(\{\mathbf{Q}_{i}\}_{i=L-r}^{L}\cdot \mathbf{K}),\quad
    \\
    \mathbf{S} &= \text{MeanPool}(\{\mathbf{W}_{i}\}_{i=1}^{r})
\end{align*}
where $\mathbf{Q}_{i}$ represents the query tensor corresponding to the $i$th vision token within the context, and $\mathbf{K}\in \mathbb{R}^{L \times d}$ denotes all cached key tensors. 
$\mathbf{W}_i\in \mathbb{R}^{r\times L}$ denotes the attention score matrix, and $\mathbf{S}\in \mathbb{R}^{L}$ denotes the pooled importance scores assigned to each vision token in the cache. 
Compared with full-scale attention computation, the aforementioned calculation incurs an additional computational and memory overhead of less than 1\%.

\begin{table}[t]
    \centering
    \caption{Comparison of performance (accuracy) on MLVU between retrieval with and without positional embeddings.}
    \vspace{-0.2cm}
    \label{tab:position_comparison}
    \resizebox{0.86\linewidth}{!}{%
    \begin{tabular}{l c c c c }
        \toprule
        \textbf{Method}  & \textbf{Holistic} & \textbf{S.D.} & \textbf{M.D.} & \textbf{All} \\
        \midrule
        LiveVLM & 82.1 & 70.3 & 39.4 & 66.2 \\
        +\textit{Retr. (w/o position)} & 83.4 & 72.3 & 41.3 & 68.1\\
        +\textit{Retr. (w/ position)} & 81.0 & 68.4 & 39.0 & 64.8\\
        \bottomrule
    \end{tabular}
    }
    \vspace{-0.4cm}
\end{table}

Based on the importance scores $\mathbf{S}$, VSB sorts all vision tokens and then evenly partitions the context length into $N$ buckets with equal capacity $B$, where $N\times B$ is equal to the cache budget $M$. Each vision token is assigned to a specific bucket according to its original position, e.g., if one token occupies the $256$th position in a context of length 1000 and the context is partitioned into 10 buckets, it shall be allocated to the 3rd bucket.
The algorithm populates the buckets through two sequential phases:
(1) In the first phase, the top-$R$ vision tokens ($R$ denotes ratio) with highest importance scores are greedily assigned to buckets, so as to ensure the retention of the most informative visual content;
(2) In the second phase, the remaining vision tokens are traversed in descending order. 
Each vision token is retained if and only if its target bucket still has available capacity. 
This phase concludes once $M$ vision tokens are selected.
Upon completion of the two phases, VSB sequentially concatenates the vision tokens within each bucket, together with their corresponding KV tensors, to construct a compressed new cache.

\begin{figure}[t]
  \centering
  \includegraphics[width=0.98\linewidth]{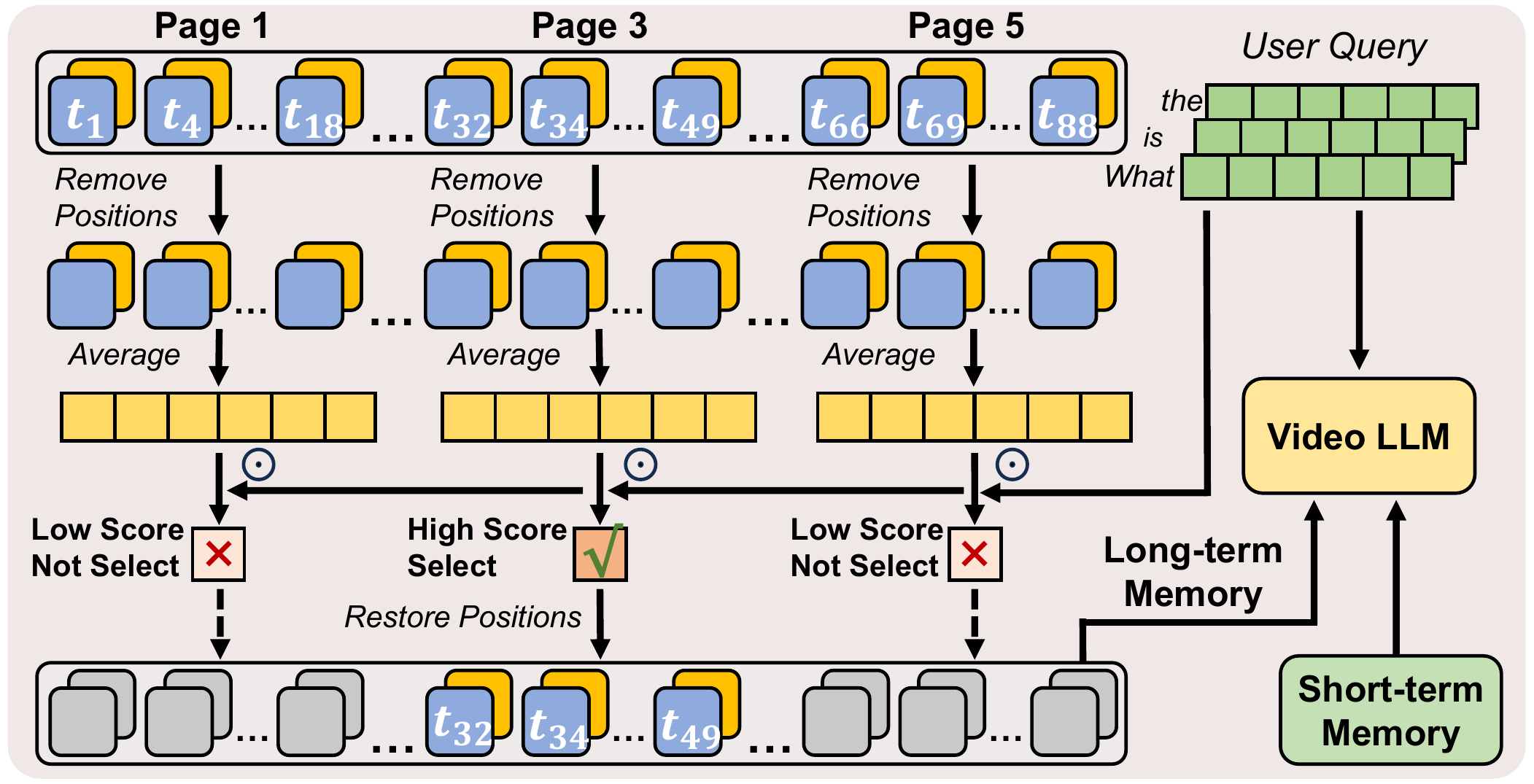}
  \vspace{-0.2cm}
  \caption{Illustration of the Position-agnostic KV Retrieval (PaR) mechanism.}
  \label{fig:PaR}
  \vspace{-1em}
\end{figure}

\section{Position-agnostic KV Retrieval}

\subsection{Discontinuous Positions Degrade Key Similarity} 

While VSB effectively facilitates the preservation of long-form video content with minimal memory overhead, its query-agnostic compression paradigm inevitably results in the retention of considerable amounts of irrelevant semantic information in the cache when the model addresses specific queries.
This phenomenon necessitates a retrieval mechanism that exclusively retrieves query-relevant KVs to support model reasoning when answering different questions based on the same streamingly updated KV cache.
Notably, existing methods \cite{quest, rekv} adopting page-level retrieval prove ineffective in compressed caches. 
The underlying reason lies in the token-discarding nature of mainstream cache compression approaches: redundant tokens are discarded, leading to discontinuous positional information for the retained tokens. 
Consequently, the key tensors corresponding to these retained tokens are augmented with discontinuous positional embeddings. 
During the retrieval process, the KVs stored in the compressed cache are required to be sequentially partitioned into multiple pages, with a representative tensor generated for each page to compute the relevance score to the query tensors of question tokens.
However, such discontinuity significantly undermines the similarity among key tensors within each page (as shown in Fig. \ref{fig:layer_cosine_similarity}), rendering it infeasible to generate a valid representative key tensor capable of characterizing the semantic information of the entire page. 
As a result, retrieval based on these representative keys leads to the introduction of misleading information and degraded response accuracy (in Tab. \ref{tab:position_comparison}).

\subsection{Decouple Retrieval from Positional Embeddings} 

To resolve this issue, we propose the Position-agnostic KV Retrieval (PaR) mechanism (as illustrated in Fig. \ref{fig:PaR}), which decouples retrieval from the discontinuous positional embeddings. 
Specifically, the mechanism removes positional embeddings integrated into key tensors during retrieval, thereby mitigating the interference caused by positional discontinuity. 
Upon acquiring position-agnostic key tensors, PaR partitions these tensors into pages and computes mean key tensors to facilitate retrieval.
Following the completion of retrieval, the corresponding positional embeddings are reintegrated into the retrieved key tensors to restore the positional information essential for model reasoning. This strategy also ensures that the retrieval depends solely on the inherent semantic information of key tensors.

During the retrieval process, PaR leverages the intrinsic attention mechanism of LLMs to evaluate the relevance between the query and video content. First, it encodes the specific question and generates corresponding query tensors using the foundation model. 
Subsequently, PaR removes the positional embeddings and partitions the KVs retained in the cache into multiple pages of equal size $C$. To generate representative tensors for each page, PaR directly averages the key tensors within the page to derive a mean key tensor, as position-agnostic key tensors exhibit high mutual similarity.
Finally, PaR concatenates all mean key tensors sequentially and performs simplified attention computation between the query tensors and the concatenated mean key tensors. 
Similar to VSB in Sec. \ref{sec:svb}, PaR takes the pooled attention scores as the metric to retrieve query-relevant KV pages. 
PaR enhances response efficiency by substantially reducing the KV cache size employed in the full-scale attention computation of the Video LLM.

\begin{table}[t]
    \centering
    \caption{Comparison of results (accuracy) on offline benchmarks. 
    The \textbf{best} and \underline{second best} results are in bold and underlined respectively.
    }
    \vspace{-0.3cm}
    \label{tab:offline}
    \resizebox{0.98\linewidth}{!}{%
    \begin{tabular}{l c c c c c c}
        \toprule
        \multirow{2}{*}{\textbf{Model}}  & \multirow{2}{*}{\textbf{Frames}} & \multirow{2}{*}{\textbf{LVB}} & \multirow{2}{*}{\textbf{MLVU}} & \multicolumn{3}{c}{\textbf{VideoMME (w/o. sub)}} \\
        \cmidrule(lr){5-7}
        & & & & Medium & Long & All \\
        \midrule
        \rowcolor{gray!20} \multicolumn{7}{c}{\textbf{Open-source Offline Video LLMs}} \\
        \midrule
        LongVA-7B & 128 & - & 56.3 & 50.4 & 46.2 & 52.6 \\
        Kangaroo-8B & 64 & 54.2 & 61.0 & 55.3 & 46.6 & 56.0 \\
        VideoXL-7B & 128 & - & 64.9 & 53.2 & 49.2 & 55.5 \\
        \midrule
        \rowcolor{gray!20} \multicolumn{7}{c}{\textbf{Open-source Online Video LLMs}} \\
        \midrule
        MovieChat-7B  & 2048 & - & 25.8 & - & 33.4 & 38.2 \\
        Dispider-7B  & 1 fps$^\dagger$ & - & 61.7 & 53.7 & 49.7 & 56.5 \\
        \midrule
        LLaVA-OneVision-7B & 32 & 55.6 & 64.7 & 54.7 & 46.2 & 
        56.9 \\
        \quad+ReKV & 0.5/0.2 fps & \underline{55.8} & \textbf{68.2} & 55.6 & 48.6 & 58.3 \\
        \quad+StreamMem & 0.5/0.2 fps & 54.4 & 66.9 & \underline{56.6} & \underline{50.1} & \underline{59.4} \\
        \rowcolor{rowgreen} \quad\textbf{+LiveVLM (ours)} & 0.5/0.2 fps & \textbf{56.1} & \underline{68.1} & \textbf{57.0}  & \textbf{51.3} & \textbf{59.6} \\
        \bottomrule
    \end{tabular}
    }
\end{table}

\begin{table}
    \centering
    \caption{Comparison of results on the RVS-Ego and RVS-Movie benchmarks.
    }
    \vspace{-0.3cm}
    \label{tab:rvs}
    \resizebox{0.98\linewidth}{!}{%
    \begin{tabular}{lcccc cc}
        \toprule
        \multirow{2}{*}{\textbf{Model}} & \multicolumn{2}{c}{\textbf{RVS-Ego}} & \multicolumn{2}{c}{\textbf{RVS-Movie}} & \multicolumn{2}{c}{\textbf{Avg.}} \\
        \cmidrule(lr){2-3} \cmidrule(lr){4-5} \cmidrule(lr){6-7}
        & Acc. & Score & Acc. & Score & Acc. & Score \\
        \midrule
        ReKV & 63.7 & 4.0 & 54.4 & 3.6 & 59.0 & 3.8 \\
        \midrule
        ReKV w/o off. & 55.8 & 3.3 & 50.8 & 3.4 & 53.3 & 3.4 \\
        Flash-VStream & 57.0 & 4.0 & 53.1 & 3.3 & 55.0 & 3.6 \\
        Infinipot-V & 57.9 & 3.5 & 51.4 & 3.5 & 54.6 & 3.5 \\
        StreamMem & 57.6 & 3.8 & 52.7 & 3.4 & 55.2 & 3.6 \\
        \rowcolor{rowgreen} \textbf{LiveVLM (ours)} & 57.8& 3.9 & 53.4 & 3.6 & \textbf{55.6} & \textbf{3.8}\\
        \bottomrule
    \end{tabular}
    }
    \vspace{-0.4cm}
\end{table}

\section{Experiments}

\subsection{Implementation Details}

\quad \textbf{Evaluation benchmarks.} We evaluate LiveVLM on both offline and online video question-answering (VideoQA) benchmarks. For offline evaluation, we adopt LongVideoBench\cite{longvideobench}, MLVU \cite{mlvu} and VideoMME \cite{videomme}. 
Specifically, we use the ``val'' dataset for LongVideo-Bench and the ``dev'' splits for MLVU, while conducting experiments on VideoMME without subtitles. 
For online evaluation, we utilize RVS-Ego, RVS-Movie \cite{vstream} and StreamingBench \cite{streamingbench}, which evaluate online video understanding capabilities from various perspectives. 


\textbf{Experimental setup.} LiveVLM is instantiated based on the foundation model \texttt{LLaVA-OneVision-Qwen2-7B-OV} \cite{llavaonevision}. 
By default, the model adopts FP16 mixed precision. 
Experiments are conducted on an NVIDIA 4090D GPU with 24 GB of memory.
The cache budget $M$ is set to 12k tokens, corresponding to a moderate memory footprint. 
For simplicity, the bucket capacity $B$ is configured to 1.
For retrieval, LiveVLM sets the page size $C$ to 16 and retrieves 40\% content from the cache.

\subsection{Main Results}
\quad \textbf{Offline video understanding.}
As shown in Table \ref{tab:offline}, LiveVLM exhibits superior overall performance in offline video understanding compared with all state-of-the-art online methods.
Notably, it excels on handling extremely long videos, yielding a 3.4\% improvement on MLVU and a 5.1\% gain on the long-form subset of VideoMME compared with the foundation model.
For the medium-form subset of VideoMME, LiveVLM also achieves the optimal score of 57.0, demonstrating robust capability across different temporal scales.

\begin{table*}
\caption{Performance comparison on StreamingBench focusing on Real-Time Visual Understanding and Omni-Source Understanding tasks. The statistics represent accuracy (\%) of question-answering. }
\vspace{-0.2cm}
\label{tab:streamingbench}
\centering
\resizebox{0.98\textwidth}{!}{%
\begin{tabular}{lc ccccc ccccc ccccc}
\toprule
\multirow{2}{*}{\textbf{Model}} & \multirow{2}{*}{\textbf{Frames}} & \multicolumn{10}{c}{\textbf{Real-Time Visual Understanding}} & \multicolumn{4}{c}{\textbf{Omni-Source Understanding}} & \multirow{2}{*}{\textbf{Overall}} \\
\cmidrule(lr){3-12} \cmidrule(lr){13-16} 
& & OP & CR & CS & ATP & EU & TR & PR & SU & ACP & CT & ER & SCU & SD & MA & \\ 

\midrule
\rowcolor{gray!20} \multicolumn{17}{c}{\textbf{Proprietary MLLMs}} \\ 
\midrule
\color{darkgraycolor}Claude 3.5 Sonnet & \color{darkgraycolor}20  & \color{darkgraycolor}73.33 & \color{darkgraycolor}80.47 & \color{darkgraycolor}84.09 & \color{darkgraycolor}82.02 & \color{darkgraycolor}75.39 & \color{darkgraycolor}79.53 & \color{darkgraycolor}61.11 & \color{darkgraycolor}61.79 & \color{darkgraycolor}69.32 & \color{darkgraycolor}43.09 & \color{darkgraycolor}31.60 & \color{darkgraycolor}34.00 & \color{darkgraycolor}32.80 & \color{darkgraycolor}48.80 & \color{darkgraycolor} 59.08 \\

\color{darkgraycolor}GPT-4o \cite{gpt-4o} & \color{darkgraycolor}64  & \color{darkgraycolor}77.11 & \color{darkgraycolor}80.47 & \color{darkgraycolor}83.91 & \color{darkgraycolor}76.47 & \color{darkgraycolor}70.19 & \color{darkgraycolor}83.80 & \color{darkgraycolor}66.67 & \color{darkgraycolor}62.19 & \color{darkgraycolor}69.12 & \color{darkgraycolor}49.22 & \color{darkgraycolor}41.20 & \color{darkgraycolor}37.20 & \color{darkgraycolor}43.60 & \color{darkgraycolor}56.00 & \color{darkgraycolor} 62.50 \\ 

\midrule
\rowcolor{gray!20} \multicolumn{17}{c}{\textbf{Open-source Offline Video LLMs}} \\ 
\midrule

LongVA-7B & 128 & 70.03 & 63.28 & 61.20 & 70.92 & 62.73 & 59.50 & 61.11 & 53.66 & 54.67 & 34.72 & 39.60 & 32.40 & 28.00 & 41.60 & 50.75 \\
Kangaroo-7B  & 64 & 71.12 & 84.38 & 70.66 & 73.20 & 67.80 & 61.68 & 56.48 & 55.69 & 62.04 & 38.86 & 37.60 & 31.20 & 28.80 & 39.20 & 53.20 \\
LLaVA-NeXT-Video-32B & 64 & 78.20 & 70.31 & 73.82 & 76.80 & 63.35 & 69.78 & 57.41 & 56.10 & 64.31 & 38.86 & 37.69 & 24.80 & 34.40 & 42.80 & 54.93 \\ 
LLaVA-OneVision-7B & 32 & 80.38 & 74.22 & 76.03 & 80.72 & 72.67 & 71.65 & 67.59 & 65.45 & 65.72 & 45.08 & 40.80 & 37.20 & 33.60 & 44.80 & 58.85 \\

\midrule
\rowcolor{gray!20} \multicolumn{17}{c}{\textbf{Open-source Online Video LLMs}} \\ 
\midrule

Flash-VStream-7B \scriptsize{[CVPR 24]}  & - & 25.89 & 43.57 & 24.91 & 23.87 & 27.33 & 13.08 & 18.52 & 25.20 & 23.87 & 48.70 & 25.91 & 24.90 & 25.60 & 28.40 & 24.28 \\ 
VideoLLM-online-8B \scriptsize{[CVPR 24]} & 2 fps & 39.07 & 40.06 & 34.49 & 31.05 & 45.96 & 32.40 & 31.48 & 34.16 & 42.49 & 27.89 & 31.20 & 26.51 & 24.10 & 32.00 & 33.15 \\ 
Disperder-7B \scriptsize{[CVPR 25]} & 1 fps & 74.92 & 75.53 & 74.10 & 73.08 & 74.44 & 59.92 & 76.14 & 62.91 & 62.16 & 45.80 & 35.46 & 25.26 & 38.57 & 43.34 & 55.65 \\ 
ReKV-7B \scriptsize{[ICLR 25]} & 0.5 fps & 74.39 & 78.91 & 78.55 & 77.12 & 68.32 & 67.91 & 67.59 & 62.60 & 64.31 & 44.56 & 38.80 & 24.80 & 39.60 & 46.40 & 57.20  \\
\midrule
\rowcolor{rowgreen} LiveVLM-7B & 0.5 fps & 79.84 & 79.69 & 84.86 & 80.72 & 67.08 & 70.09 & 74.07 & 66.26 & 67.99 & 42.49 & 49.60 & 35.20 & 48.80 & 66.00 & \textbf{63.10} \\

\bottomrule
\end{tabular}
}
\end{table*}

\textbf{Online video understanding.}
We evaluate methods via two benchmark categories: (1) RVS-Ego and RVS-Movie (generation tasks, graded by GPT-3.5-turbo-0125); (2) StreamingBench, comprising diverse Multiple-Choice Question Answering (MCQA) subtasks. 

On RVS-Ego and RVS-Movie, LiveVLM achieves the strongest performance (Tab. \ref{tab:rvs}) under 24 GB GPU memory constraint and no CPU offloading for fairness.
With 55.6\% accuracy and 3.8 score, it outperforms the leading training-free baseline StreamMem at 55.2/3.6 as well as other competing methods, including InfiniPot-V and Flash-VStream. 
While ReKV with CPU offloading reports a higher accuracy of 59.0\%, its offloading strategy incurs substantial latency, rendering it impractical for real-time streaming.

Table~\ref{tab:streamingbench} present the accuracy across two categories in StreamingBench: Real-Time Visual Understanding and Omni-Source Understanding. LiveVLM significantly outperforms SoTA online methods, improving the average accuracy by 4.25\% compared with its foundational model LLaVA-OneVision-7B. In contrast, ReKV \cite{rekv} exhibits consistent performance degradation across all tasks.
These results underscore the effectiveness of LiveVLM, especially for streaming VideoQA scenarios. 


\begin{figure}[t]
  \centering
  \includegraphics[width=0.80\linewidth]{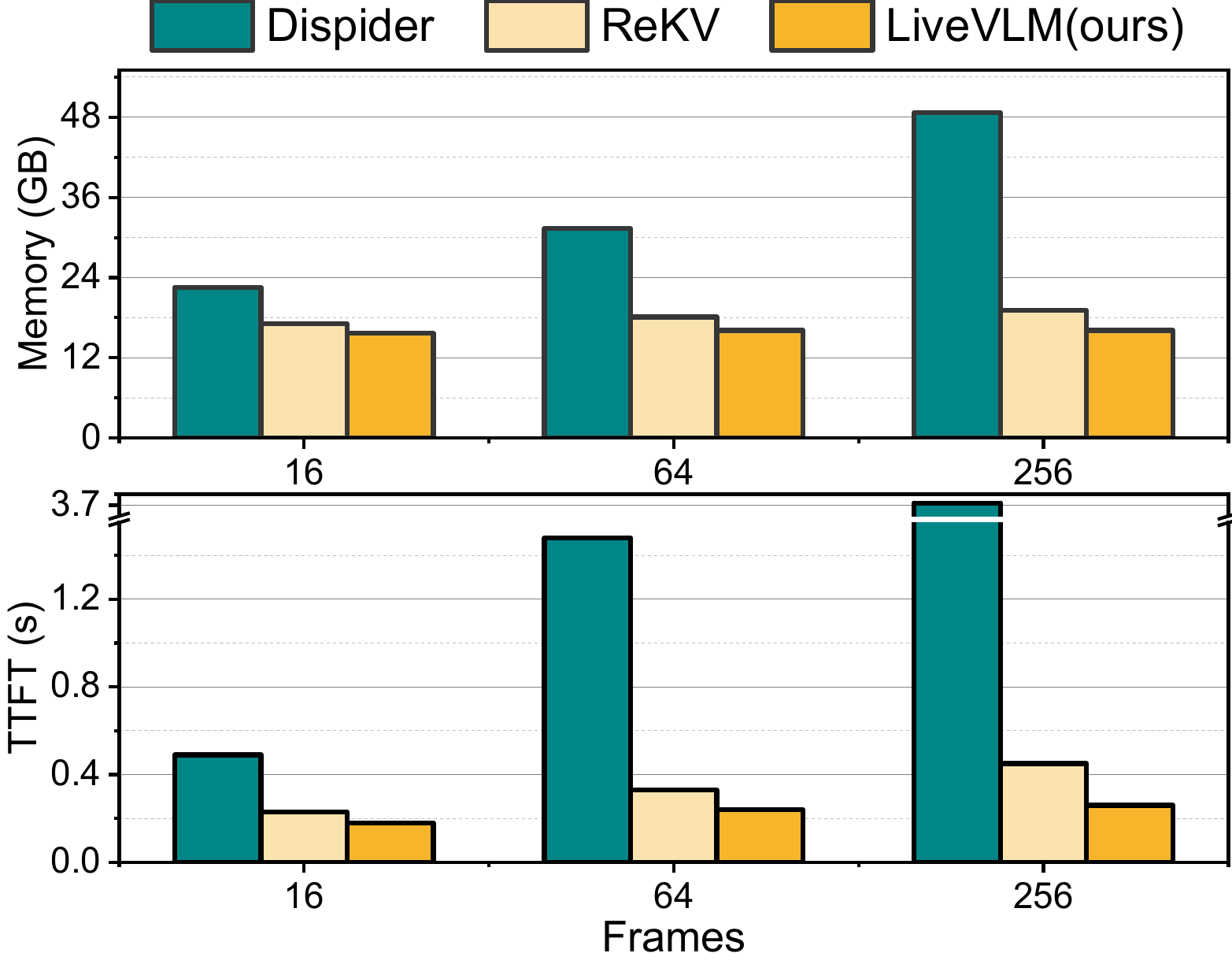}
  \vspace{-0.4cm}
  \caption{Comparison of memory and latency across varying frame budgets.}
  \label{fig:memory_speed}
  \vspace{-0.4cm}
\end{figure}

\subsection{Efficiency Analysis} 

To validate LiveVLM’s real-time efficiency, we benchmark it against SoTA online methods, including Dispider and ReKV. As shown in Fig. \ref{fig:memory_speed}, LiveVLM consistently outperforms both baselines in latency and memory usage. As input frame count increases, ReKV and LiveVLM retain stable performance, whereas Dispider’s memory usage and latency surge sharply. At 256 frames, LiveVLM reduces peak GPU memory by 3.02× (vs. Dispider) and 1.19× (vs. ReKV). In response latency, LiveVLM delivers a 1.73× speedup over ReKV, reflecting the unavoidable latency of its offloading strategy.

\subsection{Ablation Studies}

\quad \textbf{Contribution of each component.} 
To further investigate the contribution of VSB and PaR, we conduct ablation studies in Tab.\ref{tab:components}. Initially, performance on MLVU degrades significantly without any modules, as memory constraints restrict the model from sampling additional video frames for inference.
Introducing the VSB compression module enables more video frames to be streamed into the model under the same memory constraints, capturing extra visual information and improving performance. 
However, in VideoQA, cached query-agnostic visual details contain substantial irrelevant information, and direct input of all such details degrades performance. 
Therefore, integrating the PaR retrieval module with VSB further improves overall accuracy by 1.9\%. Combining both modules, our approach enhances final performance on the three subtasks by 2.6\%, 4.1\%, and 1.7\% respectively.

\begin{table}
    \centering
    \caption{Contribution of VSB and PaR.}
    \label{tab:components}
    \vspace{-0.2cm}
    \begin{tabular}{cc|ccc c}
        \toprule
        \textbf{VSB} & \textbf{PaR} & \textbf{Holistic} & \textbf{S.D.} & \textbf{M.D.} & \textbf{All} \\
        \midrule
        \scriptsize{\XSolidBrush} & \scriptsize{\XSolidBrush} & 80.8 & 68.2 & 39.6 & 64.7\\
        \Checkmark & \XSolidBrush & 82.1 & 70.3 & 39.4 & 66.2\\
        \rowcolor{rowgreen} \Checkmark & \Checkmark & \textbf{83.4} & \textbf{72.3} & \textbf{41.3} & \textbf{68.1}\\
        \bottomrule
    \end{tabular}
\end{table}

\begin{table}
    \centering
    \caption{Ablation of varying retrieval ratios.}
    \vspace{-0.2cm}
    \label{tab:retrieval_ratio}
    \resizebox{0.65\linewidth}{!}{
    \begin{tabular}{c|cccc}
        \toprule
        \textbf{Ratio} & \textbf{Holistic} & \textbf{S.D.} & \textbf{M.D.} & \textbf{All} \\
        \midrule
        0.2 & 81.0 & 70.4 & 41.3 & 66.4 \\
        \rowcolor{rowgreen}\textbf{0.4} & \textbf{83.4} & \textbf{72.3} & \textbf{41.3} & \textbf{68.1} \\
        0.6 & 82.3 & 70.8 & 40.2 & 66.7 \\
        0.8 & 81.9 & 70.0 & 40.2 & 66.2 \\
        1.0 & 82.1 & 70.3 & 39.4 & 66.2 \\
        \bottomrule
    \end{tabular}
    }
    \vspace{-0.4cm}
\end{table}

\textbf{Retrieval ratio ablation.} 
We investigate the impact of varying retrieval ratios on model performance as presented in Table \ref{tab:retrieval_ratio}. The retrieval ratio is defined as the proportion of retrieved KVs to the entire cache. As the retrieval ratio increases, the accuracy across the three subtasks initially rises and then declines. We observe that the optimal retrieval ratio is 0.4, whereas inputting the entire cache into the Video LLM results in the lowest overall accuracy. The results validate that the cache contains substantial redundancy, and our proposed retrieval method, PaR, can effectively enhance model performance.

\section{Conclusion}
In this paper, we propose LiveVLM, a training-free framework specifically designed for streaming, online video understanding and real-time interaction. LiveVLM leverages an innovative Vision Sink Bucketing mechanism to retain critical video details within a fixed-size cache, while integrating a Position-agnostic KV Retrieval mechanism to efficiently collect both short-term and long-term visual information.

\begin{acks}
    This work is sponsored by the National Natural Science Foundation of China (No. 62472273 and No. 62232015).
\end{acks}

\bibliographystyle{ACM-Reference-Format}
\bibliography{main}
\end{document}